\documentclass[letterpaper]{article} 
\usepackage[dvipsnames, svgnames, x11names]{xcolor} 
\usepackage{aaai25}  
\usepackage{times}  
\usepackage{helvet}  
\usepackage{courier}  
\usepackage[hyphens]{url}  
\usepackage{graphicx} 
\usepackage{natbib}  
\usepackage{caption} 
\usepackage{algorithm}
\usepackage{algorithmic}
\usepackage{multirow}
\usepackage{graphicx}  
\usepackage{amsmath}
\usepackage{lineno}
\usepackage{tabularx}
\usepackage{booktabs}
\usepackage{url}

\usepackage{newfloat}
\usepackage{listings}
\DeclareCaptionStyle{ruled}{labelfont=normalfont,labelsep=colon,strut=off}

\usepackage{newfloat}
\usepackage{listings}
\DeclareCaptionStyle{ruled}{labelfont=normalfont,labelsep=colon,strut=off} 
\floatstyle{ruled}
\newfloat{listing}{tb}{lst}{}
\floatname{listing}{Listing}
\title{HeightFormer: A Semantic Alignment Monocular 3D Object Detection Method from Roadside Perspective}
\author{
    Pei Liu\textsuperscript{\rm 1}\equalcontrib, 
    Zihao Zhang\textsuperscript{\rm 2}, 
    Haipeng Liu\textsuperscript{\rm 3}, 
    Nanfang Zheng\textsuperscript{\rm 4}, 
    Yiqun Li\textsuperscript{\rm 4}, 
    Meixin Zhu\textsuperscript{\rm 1}, 
    Ziyuan Pu\textsuperscript{\rm 4}
}
\affiliations{
    \textsuperscript{\rm 1}Intelligent Transportation Thrust, Systems Hub, The Hong Kong University of Science and Technology (Guangzhou), Guangzhou, 511458, China\\
    \textsuperscript{\rm 2}School of Cyber Science and Engineering, Southeast University, Nanjing, 211189, China\\
    \textsuperscript{\rm 3}Li Auto Inc, Jiading District,
Shanghai, 201800, China \\
    \textsuperscript{\rm 4}School of Transportation, Southeast University, Nanjing, 211189, China

}

\usepackage{bibentry}

\begin{document}
	
\maketitle
	
	\begin{abstract}
		The on-board 3D object detection technology has received extensive attention as a critical technology for autonomous driving, while few studies have focused on applying roadside sensors in 3D traffic object detection. Existing studies achieve the projection of 2D image features to 3D features through height estimation based on the frustum. However, they did not consider the height alignment and the extraction efficiency of bird's-eye-view features. We propose a novel 3D object detection framework integrating Spatial Former and Voxel Pooling Former to enhance 2D-to-3D projection based on height estimation. Extensive experiments were conducted using the Rope3D and DAIR-V2X-I dataset, and the results demonstrated the outperformance of the proposed algorithm in the detection of both vehicles and cyclists. These results indicate that the algorithm is robust and generalized under various detection scenarios. Improving the accuracy of 3D object detection on the roadside is conducive to building a safe and trustworthy intelligent transportation system of vehicle-road coordination and promoting the large-scale application of autonomous driving. The code and pre-trained models will be released on \textcolor{ProcessBlue}{\url{https://anonymous.4open.science/r/HeightFormer.}}
	\end{abstract}
	
	%
	
	\section{Introduction}
	Autonomous driving technology is developing rapidly as a new transportation technology paradigm for reducing traffic accidents and improving traffic efficiency. Perception technology is one of the most important technologies for autonomous driving. Autonomous driving vehicles can obtain surrounding environment information in order to make decisions and take actions by sensors \cite{van_brummelen_autonomous_2018}. According to the types of sensors adopted, perception technology can be divided into three types: point-cloud-based perception, vision-based perception, and fusion perception. Due to the high cost of lidar, many scholars believe that vision-based perception is the main research direction for promoting the mass production of autonomous vehicles in the future \cite{ma_3d_2024,muhammad_vision-based_2022}. However, on-board cameras are limited in their perception range of view due to the restricted installation height and are prone to be occluded by surrounding vehicles, especially trucks or buses. The blind spots may cause serious crashes. To fill the gap, many researchers have begun to focus on the performance of roadside cameras, expecting to provide accurate and reliable perception results for autonomous vehicles \cite{ye_yolov7-3d_2023, Hao_2024_CVPR}. 
	
	Roadside perception, due to the higher installation position of sensors, has a broader perception range and fewer blind spots compared to the perception of the ego vehicle. Based on the characteristics of fixed and higher installation positions, roadside cameras have better advantages in the 3D object detection task. In the 3D object detection task, it is not only necessary to classify objects but also to obtain information such as the position, size, orientation, and speed of the objects in the 3D space \cite{qian_3d_2022}. Therefore, many researchers have carried out research on the monocular 3D object detection task from the roadside perspective \cite{yang_monogae_2024, NEURIPS2023_2703a0e3}. However, due to the inconsistent types and installation methods of cameras on the roadside, there may be different focal lengths and pitch angles, and their perspectives are no longer parallel to the ground. The traditional perception coordinate system based on vehicles is no longer applicable to roadside perception equipment. These problems have brought many challenges to the roadside monocular 3D object detection task.
	
	The mainstream methods of vision-based 3D object detection include those leveraging the attention mechanism and those employing frustum projection at present. The method based on the attention mechanism mainly achieves detection by performing regression prediction on 3D detection boxes \cite{bevformer}. However, since cameras do not provide image depth information, the perception accuracy of this method in 3D detection is relatively low. The frustum-based method estimates the height or depth of objects in 2D images, constructs 3D objects through projection, and generates BEV (bird's eye view) features through voxel pooling to achieve object detection \cite{yang_bevheight_2023,li2023bevdepth}. Because the frustum-based method has higher accuracy, and height-based projection can improve the robustness of the algorithm. Therefore, we mainly carried out the 3D object detection task based on the idea of frustum projection. 
	
	Roadside perception can provide self-driving vehicles with more extensive and precise environmental information in the future. However, this requires first solving the noise caused by different cameras parameters and overcoming the problem of decreased object perception ability due to different pitch angles when cameras are installed, that is, to improve the robustness, accuracy and credibility of roadside perception. Based on those problems, we leveraged the height projection method of the frustum and conducted extensive experiments with the Rope3D \cite{ye2022rope3d} and DAIR-V2X-I \cite{yu_dair-v2x_2022} dataset. Compared with the previous methods, our method has achieved some improvements in the robustness and accuracy of the algorithm. The main contributions of this paper are summarized as follows:  
	
	\begin{itemize}
		\item We proposed a monocular 3D object detection method from a roadside perspective. To address the issue of spatial inconsistency when fusing the height feature with the context feature, a deformable multi-scale spatial cross-attention (DMSC) module is added in this paper to achieve spatial alignment of the height feature and background feature and improve the robustness of the algorithm.
		
		\item We added a self-attention mechanism during the BEV feature extraction process to address the issue of low accuracy in BEV feature extraction. This module takes into account global background information and, through dynamic weight adjustment, increases the efficiency of information extraction and improves the robustness during the BEV feature extraction process.
		
		\item We conducted extensive experiments to verify the accuracy of the proposed algorithm by the popular roadside perception benchmarks, Rope3D \cite{ye2022rope3d} and DAIR-V2X-I \cite{yu_dair-v2x_2022}. In terms of the detection of Cars and Big-vehicles in Rope3D, compared with the state-of-the-art (SOTA) algorithm BEVHeight++ \cite{yang_bevheight++_2023}, the accuracy has increased from 76.12\% to 78.49\% and from 50.11\% to 60.69\% respectively when the intersection over union (IoU) is 0.5. Meanwhile, by comparing the detection results under different difficulties, it has been verified that our method is robust. 
	\end{itemize}

	\section{Related Works}
	\subsection{Vision-based 3D Detection for Autonomous Driving}
\begin{figure*}[!ht] 
		\centering
		\includegraphics[width=0.95\textwidth]{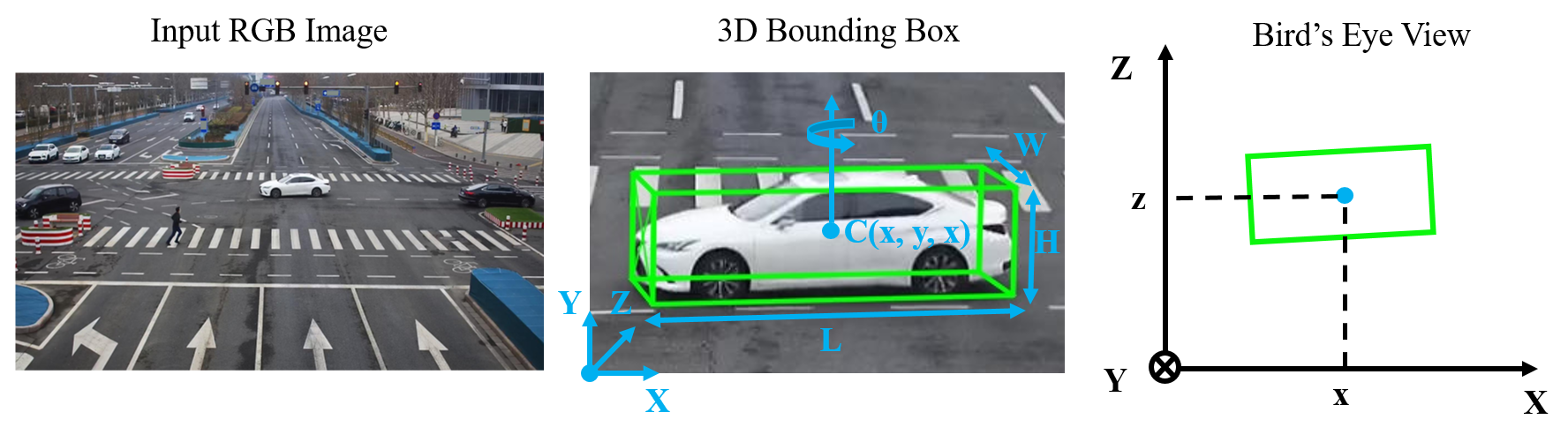}
		\caption{3D Detection Box Generation and BEV Perspective Overview Diagram. The detection box generation adopts the 7-parameter method. In the middle figure, L, W, H represent length, width, and height, respectively, C represents the coordinate (x, y, z) of the center point of the detection box, and $\theta$ represents the yaw angle.}
		\label{fig:figure1}
	\end{figure*}

	In recent years, the methods of vision-based 3D object detection have attracted the attention of both academia and industry due to their low cost and continuously improving performance \cite{ma_3d_2024}. The goal of 3D detection based on images is to output the categories and positions of objects in the input RGB images. As shown in Figure~\ref{fig:figure1}, the middle image represents the information that needs to be obtained in the 3D object detection task, which includes the center point coordinate C(x, y, z) of the object, the object's dimensions of length (L), width (W), height (H), and the yaw angle $\theta$ of the object in the Y direction. The image on the right shows the representation of the object in the BEV perspective. The BEV view can more conveniently serve the perception of downstream tasks, i.e., planning.   
	
	Currently, the image-based 3D object detection methods are mainly divided into methods transformer-based for predicting 3D detection boxes and frustum-based to estimate the depth and height of the target using 2D features. The main working principle of Transformer-based methods is to establish the connection between 3D position features and 2D image features through queries. According to different query objects, it can be divided into queries for the target set \cite{wang2022detr3d, liu2022petr, chen2022polar} and queries for the BEV grid \cite{bevformer, yang2023bevformerv2, jiang2023polarformer}. The most representative algorithms among them are DETR3D \cite{wang2022detr3d} and BEVFormer 
	\cite{bevformer}, respectively. Transformer-based methods are generally applied in multi-view vehicle-mounted scenarios \cite{caesar2020nuscenes}, and their applicability is not good for monocular roadside scenarios. 
	
	The Lift-splat-shoot (LSS) method based on frustum lifts the 2D image features into a view cone and then splats them to the BEV grid \cite{philion2020lift}. Many subsequent methods have adopted the idea of LSS \cite{huang2021bevdet, li2023bevdepth, yang_bevheight_2023, yang_bevheight++_2023}. BEVDet uses the LSS method as the view transformer to convert the image view features to the BEV space, and by constructing a unique data augmentation method and an improved non-maximum suppression (NMS), it significantly improves the performance of multi-camera 3D object detection \cite{huang2021bevdet}. Other methods adopting LSS use point cloud data as supervision to estimate the depth and height. BEVDepth is improved based on the LSS framework by introducing explicit depth supervision and a camera-aware depth estimation module and designing a depth refinement module to enhance the accuracy of depth prediction, thus achieving the new best performance in the multi-view 3D object detection task \cite{li2023bevdepth}. Since, in the roadside perspective, the optical axis of the camera is not parallel to the ground, there are many challenges in estimating the depth, and the robustness is relatively low.   
	
	Based on the features of the roadside monocular camera dataset, BEVHeight improves the LSS method by predicting the height of pixels relative to the ground instead of the depth, addressing the shortcomings of the traditional LSS method in depth prediction from the roadside perspective and enhancing the detection performance and the model's robustness to variations in camera installation height \cite{yang_bevheight_2023}. On this basis, BEVHeight++ combines depth and height estimation through the cross-attention mechanism, improving the performance of BEVHeight in the multi-view scenarios of vehicle-mounted cameras \cite{yang_bevheight++_2023}. However, when fusing the height feature and the context feature, these methods adopt a pixel-by-pixel fusion approach and do not achieve spatial alignment efficiently. Therefore, our framework efficiently fuses the height and context features through the DMSC module.  
	
	\subsection{Spatial Attention and Vision Transformer}
	The attention mechanism has been widely applied to solve the problem of key region identification in image understanding and analysis in computer vision, such as image classification, object detection, semantic segmentation, video understanding, image generation, 3D vision, multimodal tasks, and self-supervised learning \cite{guo2022attention}. The Vision Transformer (ViT) splits the image into multiple patches and regards these patches as elements in a sequence. Then, it processes the sequence of patches using a standard Transformer encoder like natural language processing \cite{dosovitskiy2020vit}. After pre-training on large-scale datasets, it demonstrates outstanding performance on various image recognition tasks. Subsequently, Xu et al. innovated the heterogeneous multi-agent self-attention and multi-scale window self-attention modules based on ViT, effectively improving the 3D object detection performance of the algorithm in complex noisy environments \cite{xu2022v2xvit}. The development of the ViT framework has significantly enhanced the performance of computer vision algorithms in tasks such as image classification and 3D object detection. 
	
	Besides ViT, spatial attention \cite{carion2020end} and temporal attention \cite{xu2017jointly} have been widely applied to computer vision tasks to explore global spatial and temporal information and avoid limited perceptual fields from convolution. VISTA enhances the 3D object detection performance of the algorithm through the cross-view spatial attention module, enabling the attention mechanism to focus on specific target areas \cite{deng2022vista}. To address the problems of slow convergence speed and limited spatial resolution existing in the spatial attention mechanism, Deformable-DETR enables the attention module to focus only on a small number of key sampling points around the reference points \cite{zhu2020deformable}. Considering that we need to match the height features with the background features, the deformable multi-scale spatial cross-spatial attention module is adopted to efficiently fuse the height feature and the context features and to achieve the improvement of the algorithm performance. Meanwhile, it enhances the efficiency of BEV feature extraction through ViT to improve the accuracy of 3D object detection.

	\section{Methodology}
\begin{figure*}[!ht]
		\centering
		\includegraphics[width=0.95\textwidth]{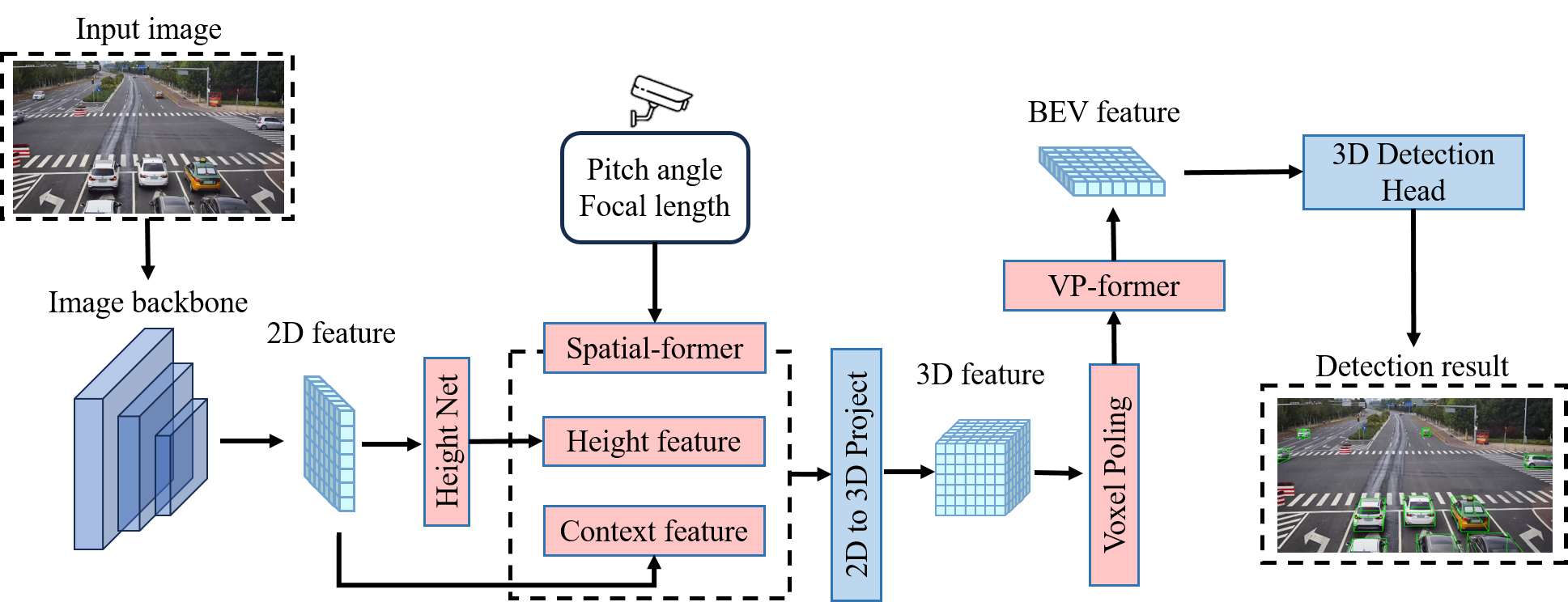}
		\caption{Overview of Our Method Architecture. The left-top is the input image; the image backbone extracted 2D features from the input image. After fusing the height features, context features and camera parameters, projected these features into 3D features by projector. Then, BEV features can be obtained by voxel pooling and the self-attention module. Finally, the 3D object detection head obtained detected results.}\label{fig:figure2}
	\end{figure*}

	Figure~\ref{fig:figure2} shows the overall framework of the proposed roadside monocular 3D object detection algorithm. Initially, an image of a fixed size is input, and its 2D features are extracted through the backbone network. Subsequently, the height features of the image are obtained through the height network \cite{yang_bevheight_2023} and combined with the context features and camera parameters to achieve the fused features considering the camera parameters and the object height features. The improved project method in 2D to 3D projector is adopted to extract 3D objects, and then voxel pooling is performed on the 3D features. The BEV features are obtained by the self-attention module, ultimately yielding the results of 3D objects through the detection head. 
	
	\subsection{Fusion of Height Feature with Context}
	As shown in Figure~\ref{fig:figure2}, height information is a critical feature. The Height Net extracts the height information from the 2D image features, and the output height feature map is the same shape as the input image. Since the traditional LSS perception method would lose the height information during the voxel pooling process \cite{yang_bevheight++_2023}, it is necessary to fuse the height feature with the environmental feature before implementing the lift operation to retain the height information. Existing methods adopt the pixel-by-pixel stitching approach to fuse the height and background information \cite{yang_bevheight_2023}. However, in roadside cameras, spatial misalignment may occur due to lighting conditions and viewpoint angles. Therefore, we adopted the DMSC module to fuse the height and context features before "lifting" as shown in the following equation:
	\begin{linenomath}
		\begin{flalign}
			F_{fused}=DMSC(F_{context},\mathrm{ }H_{pred})\mathrm{ }
		\end{flalign}
	\end{linenomath}
	Where $F_{fused}$ is the fused feature, $DMSC\left( \cdot \right) $ is the deformable multi-scale spatial cross-attention module, $F_{context}$ denotes the context feature, $H_{pred}$ is the height feature obtained by the height network prediction.
	
\begin{figure}[!ht] 
		\centering
		\includegraphics[width=0.5\textwidth]{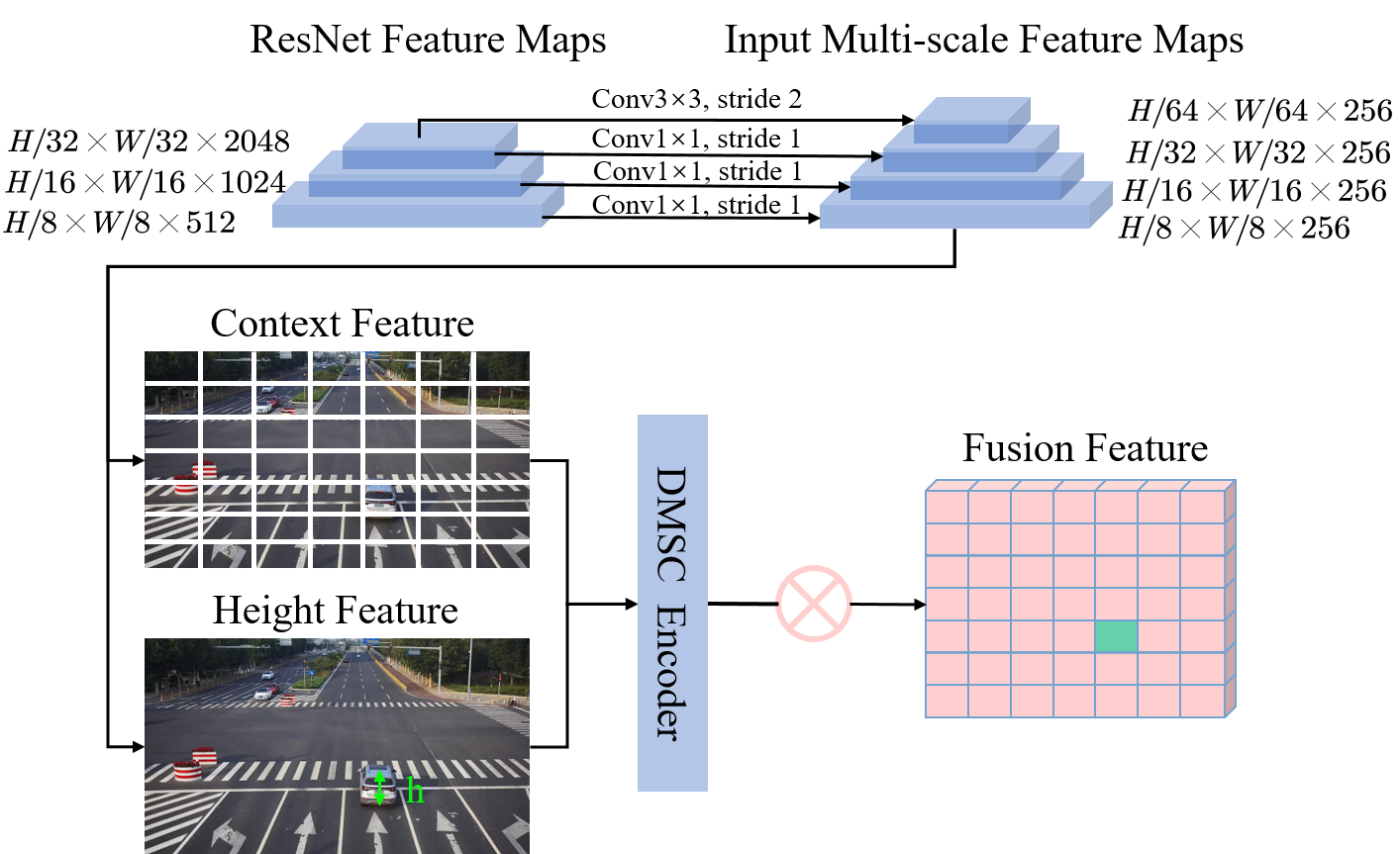}
		\caption{Deformable Multi-scale Spatial Cross-attention Fused Height Feature and Context Feature.}\label{fig:figure3}
	\end{figure}

	Meanwhile, Figure~\ref{fig:figure3} shows the detailed process of the fusion of height features and context features. Based on the 2D features of the image, the height of the traffic target in the image is estimated through the height network, and the estimated height features and the context features are used together as the input of the DMSC module. Through the DMSC module, the 2D fused features that fuse the height and background features are obtained.  
	
	\subsection{2D to 3D Projection Based on Frustum}
\begin{figure}[!ht] 
		\centering
		\includegraphics[width=0.48\textwidth]{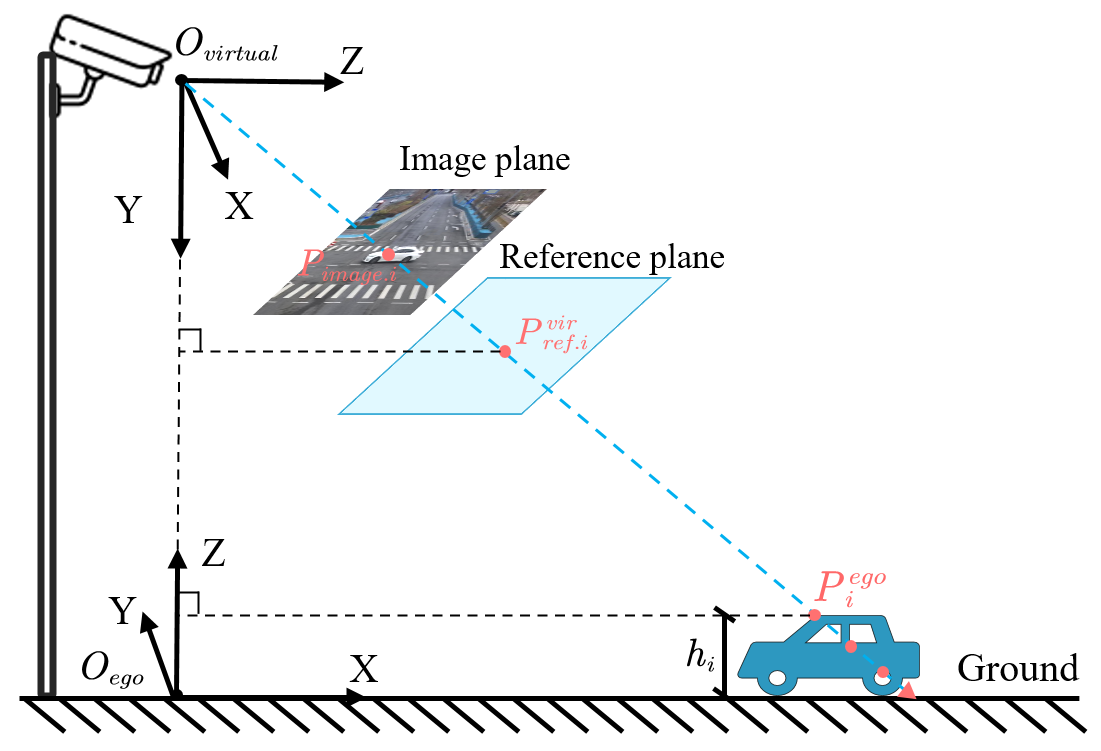}
		\caption{The Diagram Illustrating the Frustum Projection Using Height Estimation. The bounding box features detected by 2D are fused into the viewpoint and projected into the 3D space through height estimation.}\label{fig:figure4}
	\end{figure}

	We adopted the frustum projection method \cite{yang_bevheight_2023}, which is suitable for height estimation for projecting the 2D features into the 3D space. The specific projection process is shown in Figure~\ref{fig:figure4}. Since the 2D features incorporate the height information, for any pixel point $P_{image.i}$ in the image, it can be represented by the pixel position $(u, v)$ and the height $h_i$, that is, $P_{image.i} = (u, v, h_i)$. Firstly, construct a coordinate system with the virtual coordinate origin $O_{virtual}$ and a reference plane with a depth of 1, and convert the pixel points into the points in the camera coordinate system according to the internal parameter matrix of the camera. It is as follows:  
	\begin{linenomath}
		\begin{equation}
			P_{ref.i}^{cam}=K^{-1}\left[ u, v, 1, h_i \right] ^T
		\end{equation}
	\end{linenomath}
	where $P_{ref.i}^{cam}$ represents the $i$-th point in the camera coordinate system of the reference plane, and $K$ represents the internal parameter matrix of the camera.  
	
	Since the camera coordinate is generally not perpendicular to the ground, it needs to be converted to a virtual coordinate system where the Y-axis is perpendicular to the ground. 
	\begin{linenomath}
		\begin{equation}
			P_{ref.i}^{virt}=T_{cam}^{virt}P_{ref.i}^{cam}
		\end{equation}
	\end{linenomath}
	Where $P_{ref.i}^{virt}$ is the $i$-th point in the virtual coordinate system of the reference plane, and $T_{cam}^{virt}$ is the transformation matrix from the camera coordinate system to the virtual coordinate system. 
	
	Suppose the distance from the origin of the virtual coordinate system to the ground is $H$. The point $P_{gd.i}^{virt}$ in the virtual coordinate corresponding to the estimated height $h_i$ in the ground plane is found through the principle of similar triangles, as follows:  
	\begin{linenomath}
		\begin{equation}
			P_{gd.i}^{virt}\,\,=\,\,\frac{H-h_i}{y_{ref.i}^{virt}}P_{ref.i}^{virt}
		\end{equation}
	\end{linenomath}
	Where $y_{ref.i}^{virt}$ represents the coordinate of point $P_{ref.i}^{virt}$ in the Y-axis direction.  
	
	After a series of coordinate transformations and plane transformations, the point $P_{gd.i}^{virt}$ in the virtual coordinate system on the ground plane is obtained. In order to facilitate the detection of its position, it needs to be converted to the vehicle coordinate system as follows:  
	\begin{linenomath}
		\begin{equation}
			P_{gd.i}^{ego}=T_{virt}^{ego}P_{gd.i}^{virt}
		\end{equation}
	\end{linenomath}
	
	\subsection{BEV Feature Extraction}
	Compared with the 3D feature space, the BEV feature space can better capture the positions of traffic objects, thereby generating 3D detection boxes. Meanwhile, pooling the voxel features into BEV features reduces the computational load and is more adaptable to the computing characteristics of roadside perception devices. In order to improve the extraction efficiency of BEV features, we processed the pooling feature map using the self-attention module to reduce the noise interference of voxel pooling. The pooling operation is as follows:  
	\begin{linenomath}
		\begin{equation}
			F_{BEV}=VPF(F_{Pooling})\mathrm{ }
		\end{equation}
	\end{linenomath}
	Where $F_{BEV}$ represents the self-attention module, $VPF\left( \cdot \right) $ is the voxel poling former module, and $F_{Pooling}$ means the feature layer after the voxel pooling.
	
\begin{figure}[!ht] 
		\centering
		\includegraphics[width=0.48\textwidth]{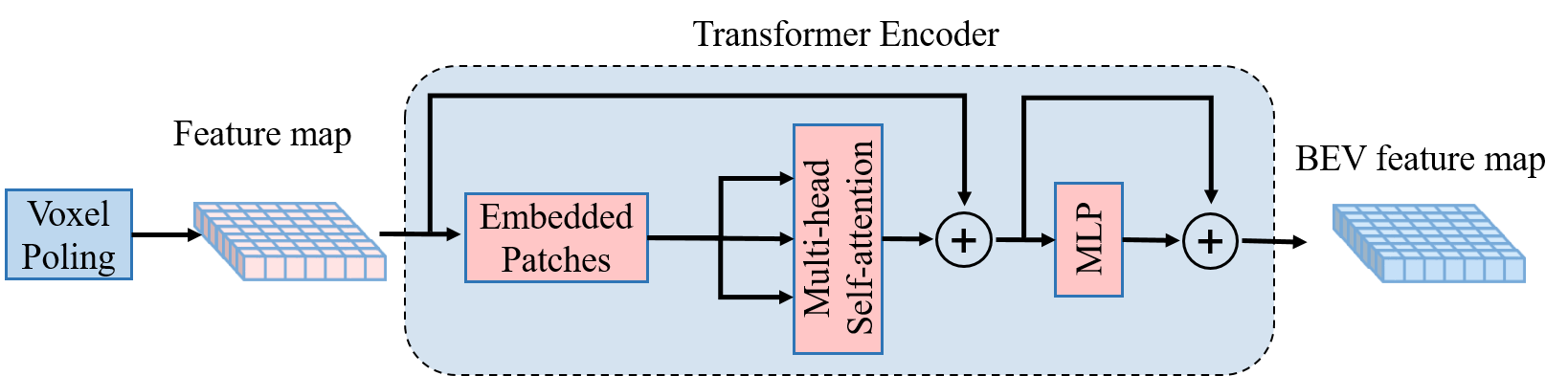}
		\caption{Schematic Diagram of Obtaining BEV Features by Self-Attention. The input on the left is the feature map after voxel pooling. The feature map is matched with fixed-size patches, and the BEV feature map is obtained through the multi-head attention mechanism and the MLP module.}\label{fig:figure5}
	\end{figure}

	The details of the method are shown in Figure~\ref{fig:figure5}: For the feature map, set an appropriate patch size and divide the features into multiple patches. The smaller the patch size, the finer the division, and the more feature details are obtained. After the division is completed, the features are processed through the self-attention heads. Each head independently calculates the weights between different patches, and the features are further processed and refined through the multi-layer perceptron (MLP) module. Finally, the features are aggregated to obtain the BEV feature map.  
	
	\section{Experiments}
	\subsection{Dataset}
	\textbf{Rope3D.}The data adopted in this experiment is the roadside camera dataset Rope3D, which was launched by the Institute for AI Industry Research (AIR), Tsinghua University in 2022 \cite{ye2022rope3d}. This dataset collected about 50,000 images at 26 intersections under different time periods, weather conditions, densities, and distributions of traffic participants through roadside cameras. It has rich and diverse image data. Meanwhile, as the positions and pitch angles where the cameras are installed are different, and the focal lengths of different cameras are also different, these differences raise the requirements for the robustness of the detection algorithm. It is worth mentioning that this dataset acquired point cloud data through vehicles equipped with LiDAR to calibrate the 3D information of traffic targets in the image data in the same scenarios as the camera collection. So, Rope3D has a relatively high annotation quality. This dataset annotates about 1,500,000 3D traffic objects. The marked traffic participants in this dataset include four major categories: Car, Big-vehicle, Pedestrian, and Cyclist. It is to be subdivided into 13 minor categories, such as Big-vehicle, which includes Truck and Bus. The label distribution of various categories is relatively uniform, reducing the influence of the long-tail effect on the detection tasks and algorithms. As a popular roadside perception benchmark, in order to uniformly compare the performance of various algorithms, this dataset proposed $AP_{3D|R40}$ and $Rope_{score}$ as the evaluation indicators of the data. At present, some researchers have tested 3D object detection algorithms in the monocular roadside view based on this dataset, such as YOLOv7-3D \cite{ye_yolov7-3d_2023}, BEVhehight \cite{yang_bevheight_2023}, BEVSpread \cite{wang_bevspread_2024}.
	
	\textbf{DAIR-V2X-I.} In order to promote the computer vision research and innovation of Vehicle-Infrastructure Cooperative Autonomous Driving (VICAD), Yu et al. released the DAIR-V2X dataset \cite{yu_dair-v2x_2022}. This is the first large-scale, multimodal, and multi-view real vehicle scene dataset, containing 71,254 frames of LiDAR (Light Detection and Ranging) frames and the same number of camera frames. All frames are from real scenes and come with 3D annotations. In this paper, we mainly use the dataset DAIR-V2X-I for the roadside view among them. This dataset contains 10,000 images, which are divided into training set, validation set and test set in the ratio of 5:2:3. Since the test set data has not been released yet, we use the validation set to verify the performance of the model. The main verification metric adopted is average perception.

	\subsection{Experiment Setting}
	To improve the robustness of the 3D object detection algorithm and enhance the credibility of the roadside perception system, we conducted extensive experiments based on the Rope3D and DAIR-V2X-I dataset. For the architectural details of the experiment, we adopt ResNet-101 \cite{he_resnet_2016} as the backbone network for image feature extraction and set the resolution of the input image as (864, 1536). Meanwhile, in order to increase the robustness of the algorithm, we adopted the method of data augmentation, that is, scaling and rotating the original image data. The experimental equipment we adopted is six NVIDIA A800 GPUs for the training and validation of the model. The data batch on each GPU device is set to 2, and we employ the AdamW optimizer \cite{kingma2014adam} with an initial learning rate of $2e-4$. In order to compare with other algorithms, the number of training epochs is 150 \cite{yang_bevheight++_2023}.  
	
	\subsection{Metrics}
	In order to compare with other State-of-the-art (SOTA) algorithms, we referred to the metric $AP|_{R_{40}}$ \cite{simonelli_disentangling_2019} used in the Rope3D dataset and $AP$ used in DAIR-V2X-I dataset as the main evaluation metric to evaluate the accuracy of the model algorithm. The calculation method of the metric is as follows: 
	\begin{linenomath}
		\begin{equation}
			\label{equation7} 
			AP|_R=\frac{1}{\left| R \right|} \sum_{r\in R}{\underset{\tilde{r} : \tilde{r} \ge  r} {\max}\rho \left( \tilde{r} \right)}
		\end{equation}
	\end{linenomath}
	Where $\rho \left( \tilde{r} \right) $ represents the accuracy rate under the determined regression threshold $r\in$$\{$1/40, 2/40,...,1$\}$.
	
	Meanwhile, an indicator that is more suitable for the Rope3D dataset is also adopted \cite{ye2022rope3d}. This indicator incorporates Average Ground Center Similarity (ACS), Average Orientation Similarity (AOS), Average Four Ground Points Distance (AGD), and Average Four Ground Points Similarity (AGS) by setting weights, assume $S=(ACS+AOS+AAS+AGS)/4$. As shown in the following equation: 
	\begin{linenomath}
		\begin{equation}
			Rope_{score}=\left( \omega _1*AP\,\,+\,\,w_2*S  \right)/ \left( \omega_1 + \omega_2 \right)
		\end{equation}
	\end{linenomath}
	Where AP is the value calculated by equation~\ref{equation7}, the value of the weight $\omega_1$ is set to 8, and the value of $\omega_2$ is set to 2.  
	
	\subsection{Comparing with State-of-the-art}
	\textbf{Results on the Rope3D.} We mainly compared 3D object detection methods that adopt the BEV method, including BEVFormer \cite{bevformer}, BEVDepth \cite{li2023bevdepth}, BEVHeight \cite{yang_bevheight_2023}, and BEVHeight++ \cite{yang_bevheight++_2023}. Among them, BEVHeight mainly focuses on roadside monocular detection, BEVFormer and BEVDepth focus on multi-view 3D object detection, and BEVHeight++ integrates height and depth information and also has a good performance in multi-view scenarios. Table~\ref{tab:table1} shows the detection situations of various IoU for Car and Big-vehicle under the Rope3D test dataset. The results show that compared with the SOTA algorithm BEVHeight++, our algorithm increases the detection accuracy of Car and Big-vehicle by 2.37\% and 10.58\%, respectively, when IoU is 0.5, and increases the detection accuracy of Big-vehicles by 1.40\% when IoU is 0.7. However, the detection result of small cars lags behind the BEVHeight++ algorithm by 4.84\%. 
	\begin{table*}[!ht]
		\centering
		\scalebox{1.0}{
			\begin{tabular}{c|cccc|cccc}
				\hline
				\multirow{3}{*}{Method} & \multicolumn{4}{c|}{IoU=0.5}                                           & \multicolumn{4}{c}{IoU=0.7}                                           \\ \cline{2-9} 
				& \multicolumn{2}{c|}{Car}            & \multicolumn{2}{c|}{Big-vehicle} & \multicolumn{2}{c|}{Car}            & \multicolumn{2}{c}{Big-vehicle} \\
				& $AP_{3D}$  & \multicolumn{1}{c|}{$R_{score}$} & $AP_{3D}$           & $R_{score}$          & $AP_{3D}$  & \multicolumn{1}{c|}{$R_{score}$} & $AP_{3D}$           & $R_{score}$        \\ \hline
				BEVFormer               & 50.62 & \multicolumn{1}{c|}{58.78}  & 34.58          & 45.16           & 24.64 & \multicolumn{1}{c|}{38.71}  & 10.08          & 26.16          \\
				BEVDepth                & 69.63 & \multicolumn{1}{c|}{74.70}  & 45.02          & 54.64           & 42.56 & \multicolumn{1}{c|}{53.05}  & 21.47          & 35.82          \\
				BEVHeight               & 74.60 & \multicolumn{1}{c|}{78.72}  & 48.93          & 57.70           & 45.73 & \multicolumn{1}{c|}{55.62}  & 23.07          & 37.04          \\
				BEVHeight++             & 76.12 & \multicolumn{1}{c|}{80.91}  & 50.11          & 59.92           & \textbf{47.03} & \multicolumn{1}{c|}{\textbf{57.77}}  & 24.43          & \textbf{39.57}          \\
				Ours                    & \textbf{78.49}  & \multicolumn{1}{c|}{ \textbf{81.72}}       &\textbf{60.69}                &\textbf{67.15}                 &42.19      & \multicolumn{1}{c|}{52.70}       &\textbf{25.83 }               &39.33                \\ \hline
				\multicolumn{9}{l}{AP and Rope represent $AP|_{R_{40}}$ and $Rope_{score}$} 
		\end{tabular}}
		\caption{Vision-based 3D Detection Performance of Car and Big-vehicle on the Rope3D Val set.}
		\label{tab:table1}
	\end{table*}  
	
	To test the robustness of the algorithm, we divided the difficulty level of object detection into three grades: Easy, Mid, and Hard, according to the distance between the object and the camera, occlusion, and truncation. Easy corresponds to objects without occlusion and no occlusion segments, Mid corresponds to targets with lateral occlusion at 0\% - 50\%, and Hard corresponds to targets with longitudinal occlusion at 50\% - 100\%. Then, the detection accuracy of the three traffic objects, Vehicle, and Cyclist, was compared. Considering that the objects of Cyclist are relatively small and the detection is more difficult, referring to the existing methods \cite{yang_bevheight_2023, yang_bevheight++_2023}, the IoU was set to 0.25 in this paper. Table~\ref{tab:table2} indicates that for the two detection targets of Vehicle and Cyclist under different difficulty conditions, our algorithm has improved by 8.71\%/9.41\%/9.23\% and 6.70\%/5.28\%/4.71\% respectively compared to the SOTA algorithm BEVheight++. According to the performance decline under different difficulties, our algorithm shows better robustness. 
	
	
	\begin{table}[!ht]
		\centering
		\scalebox{0.8}{
			\begin{tabular}{cccc|ccc}
				\hline
				\multicolumn{1}{c|}{\multirow{2}{*}{Method}} & \multicolumn{3}{c|}{Veh. (IoU=0.5)}               & \multicolumn{3}{c}{Cyc. (IoU=0.25)}              \\
				\multicolumn{1}{c|}{}                        & Easy           & Mid            & Hard           & Easy           & Mid            & Hard           \\ \hline
				\multicolumn{1}{c|}{BEVFormer}               & 61.37          & 50.73          & 50.73          & 22.16          & 22.13          & 22.00          \\
				\multicolumn{1}{c|}{BEVDepth}                & 71.56          & 60.75          & 60.85          & 40.83          & 40.66          & 40.26          \\
				\multicolumn{1}{c|}{BEVHeight}               & 75.58          & 63.49          & 63.59          & 47.97          & 47.45          & 48.12          \\
				\multicolumn{1}{c|}{BEVHeight++}             & 76.98          & 65.52          & 65.64          & 48.14          & 48.11          & 48.63          \\
				\multicolumn{1}{c|}{Ours}                    & \textbf{85.69} & \textbf{74.93} & \textbf{74.87} & \textbf{54.84} & \textbf{53.39} & \textbf{53.34} \\ \hline
				\multicolumn{7}{l}{Veh., Cyc. represent Vehicle and Cyclist respectively.}                                                                         \\
				\multicolumn{7}{l}{Easy, Mid and Hard indicate the difficulty setting levels of detection.}                                                       
		\end{tabular}}
		\caption{vision-based 3D detection performance of Car and Cyclist on the Rope3D val set.}
		\label{tab:table2}
	\end{table}

\begin{figure}[!ht]  
		\centering
		\includegraphics[width=0.48\textwidth]{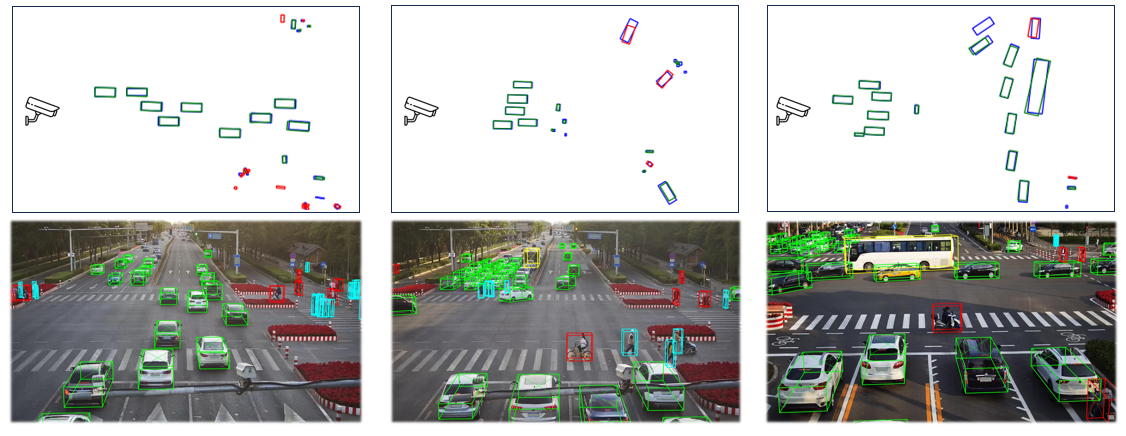}
		\caption{Visualized Detection Results. In the upper chart, the camera icon indicates the direction in which the camera is located. Blue represents the labeled detection box, green indicates correct detection, and red indicates incorrect detection, with no marking for undetected items. In the lower chart, the green detection box represents a small car, the yellow detection box represents a big-vehicle and the red detection box represents a cyclist.
		}
		\label{fig:figure7}
	\end{figure}

	Figure~\ref{fig:figure7} shows the detection results of traffic targets. This indicates that the algorithm has a higher detection accuracy for larger traffic targets located in the center of the image. However, for smaller traffic targets on both sides of the image, the effect is not ideal. 
	
	\textbf{Results on the DAIR-V2X-I.} For the validation results on the DAIR-V2X-I dataset, we compared multiple SOTA algorithms such as PointPillars \cite{lang2019pointpillars}, SECOND \cite{yan2018second}, MVXNet \cite{sindagi2019mvx}, ImvoxclNet\cite{rukhovich2022imvoxelnet}, BEVFormer \cite{bevformer}, BEVDepth \cite{li2023bevdepth}, BEVHeight \cite{yang_bevheight_2023}, BEVHeight++ \cite{yang_bevheight++_2023}, etc. As can be seen from Table~\ref{tab:dair_sota_2}, our algorithm is comprehensively superior to the existing algorithms in the 3D object detection tasks of vehicles and cyclists, and is superior to other vision-based algorithms in the hard mode of pedestrian detection. Compared with the height-estimation-based benchmark algorithm BEVHeight, our algorithm improves by 1.65\%/3.44\%/3.37\% in the Vehicle detection task, by 1.49\%/1.55\%/1.67\% in the pedestrian detection task, and by 0.56\%/0.57\%/0.59\% in the cyclist detection task.   
	
	\begin{table}[!ht]
		\scriptsize\centering\addtolength{\tabcolsep}{-3.8pt}
		\begin{tabularx}{1.0\linewidth}{l|c|ccc|ccc|ccc}
			\hline
			\multirow{2}{*}{Method} &  \multirow{2}{*}{M} &  \multicolumn{3}{c|}{$\text{Veh.}_{(IoU=0.5)}$} & \multicolumn{3}{c|}{$\text{Ped.}_{(IoU=0.25)}$} & \multicolumn{3}{c}{$\text{Cyc.}_{(IoU=0.25)}$} \\  
			~ & ~  & Easy & Mid & Hard & Easy & Mid & Hard & Easy & Mid & Hard  \\
			\hline
			PointPillars & L &63.07 & 54.00 & 54.01 & 38.53 & 37.20 & 37.28 & 38.46 & 22.60 & 22.49 \\
			SECOND & L &71.47 & 53.99 & 54.00 & 55.16 & 52.49 & 52.52 & 54.68 & 31.05 & 31.19 \\
			MVXNet & LC &71.04 & 53.71 & 53.76 & 55.83 & 54.45 & 54.40 & 54.05 & 30.79 & 31.06 \\
			\midrule
			ImvoxelNet &C & 44.78 & 37.58 & 37.55 & 6.81 & 6.746 & 6.73 & 21.06 & 13.57 & 13.17 \\
			BEVFormer & C 	&	61.37&	50.73&	50.73&	16.89&	15.82&	15.95	&22.16&	22.13&	22.06\\
			BEVDepth&	C 	&	
			75.50&	63.58&	63.67&	34.95&	33.42&	33.27& 55.67&	55.47&	55.34\\
			\midrule
			BEVHeight & C &	
			77.78&	65.77&	65.85&	41.22&	39.29&	39.46	&60.23&	60.08&	60.54\\
			BEVHeight++ & C &	
			79.31&	68.62&	68.68&	\textbf{42.87}&	\textbf{40.88}&	41.06	&60.76&	60.52&	61.01\\
			Ours & C &	
			\textbf{79.43}&	\textbf{69.21}&	\textbf{69.22}&	{42.79}&	{40.84}& \textbf{41.13}	& \textbf{60.79} &	\textbf{60.65}&	\textbf{61.13}\\
			\hline
			\multicolumn{8}{l}{\scriptsize{M, L, C denotes modality, LiDAR, camera respectively.}}
		\end{tabularx}
		\caption{Comparing with the state-of-the-art on the DAIR-V2X-I val set.}
		\label{tab:dair_sota_2}
	\end{table}

	\section{Discussion}
	Aiming at the problems existing in roadside perception, we added the Spatial Former and Voxel Pooling Former modules on the basis of the existing height estimation method framework to improve the spatial alignment ability of height features and context features and the extraction efficiency of pooling features. To verify the effectiveness of our method, extensive experiments have been conducted based on the public dataset Rope3D \cite{ye2022rope3d} and DAIR-V2X-I \cite{yu_dair-v2x_2022}. The experimental results show that the roadside monocular 3D object detection framework we proposed has outperformed the currently popular algorithms in the detection effect of vehicles and cyclists. Meanwhile, the results of different detection difficulties indicate that our algorithm is robust and can still detect traffic targets well as the detection difficulty increases. Currently, many countries around the world have installed roadside cameras. Applying these devices to the development of autonomous driving technology will provide a safer and more reliable autonomous driving environment and promote the development of intelligent transportation \cite{z_zou_real-time_2022, cres_intelligent_2024}. Our algorithm can provide technical support for the large-scale application and implementation of autonomous driving and promote the development of an intelligent transportation system of vehicle-road coordination. 
	
	Although the algorithm we proposed has achieved good results on vehicles and cyclists, there are still the following some deficiencies, and the subsequent research will continue to be improved. Firstly, since pedestrians usually appear on both sides of the road and it is difficult for roadside cameras to clearly capture pedestrian information, we did not detect pedestrian targets. Subsequently, the model framework will be enhanced based on the characteristics of pedestrian targets from the roadside perspective to achieve 3D detection of pedestrians at intersections. Finally, due to limited computing resources, we did not conduct ablation experiments to fully demonstrate the rationality of the model structure. In the subsequent research, a large number of ablation experiments will be conducted to compare the effects of different modules. 
	
	\section{Conclusion}
	In the field of 3D object detection for autonomous driving, there are problems such as low perception accuracy and weak algorithm robustness caused by equipment specifications, installation angles, and non-parallelism between the camera optical axis and the ground in roadside monocular detection algorithms. We proposed a roadside monocular 3D object detection framework based on Spatial Former and Voxel Pooling Former, which improves the performance and robustness of the algorithm based on the height estimation method. Through extensive experiments on the popular Rope3D and DAIR-V2X-I benchmark, the results show that in the 3D object detection of Big-vehicle when IoU is 0.5, our algorithm has increased from 50.11\% to 60.69\% compared to the SOTA algorithm BEVHeight++. In the detection of vehicles and cyclist under different detection difficulties, it has improved by 8.71\%/9.41\%/9.23\% and 6.70\%/5.28\%/4.71\%, respectively. The experimental results of DAIR-V2X-I show that the algorithm proposed in this paper improves by 1.65\%/3.44\%/3.37\%, 1.49\%/1.55\%/1.67\%, and 0.56\%/0.57\%/0.59\% respectively in the detection of vehicles, pedestrians, and cyclist compared to the height-estimation-based benchmark algorithm BEVHeight. Experimental results show that our algorithm has good detection performance for vehicles and cyclists. At the same time, as the detection difficulty increases, the robustness of the algorithm is good. Our algorithm improves the performance of the roadside monocular 3D object detection method, enhances the safety and reliability of autonomous driving perception, helps build an intelligent transportation system of vehicle-road coordination, and promotes the application of autonomous driving technology.

	

	
	\bibliography{aaai25}
	
\end{document}